\documentclass[conference]{IEEEtran}
\IEEEoverridecommandlockouts 

\usepackage{amsmath,amssymb,amsfonts}
\usepackage{algorithmic}
\usepackage{graphicx}
\usepackage{textcomp}
\usepackage{xcolor}
\usepackage{booktabs}
\usepackage{makecell}
\usepackage{multirow}
\usepackage{tabularray}
\usepackage[numbers, sort]{natbib} 
\usepackage{hyperref}

\def\BibTeX{{\rm B\kern-.05em{\sc i\kern-.025em b}\kern-.08em
    T\kern-.1667em\lower.7ex\hbox{E}\kern-.125emX}}
\begin{document}

\newcommand{\minisection}[1]{\vspace{0.05in} \noindent {\bf #1} \ }

\title{A Benchmark Environment for Offline Reinforcement Learning in Racing Games}

\IEEEoverridecommandlockouts
\IEEEpubid{\makebox[\columnwidth]{ 979-8-3503-5067-8/24/\$31.00~\copyright2024 IEEE \hfill} 
\hspace{\columnsep}\makebox[\columnwidth]{ }}

\author{
Girolamo Macaluso$^{1}$,
Alessandro Sestini$^{2}$, and
Andrew D. Bagdanov$^{1}$
\\
$^{1}$\textit{University of Florence}, $^{2}$\textit{SEED - Electronic Arts (EA)}
\\
\{girolamo.macaluso, andrew.bagdanov\}@unifi.it, asestini@ea.com
}

\maketitle 

\IEEEpubidadjcol
\begin{abstract}

Offline Reinforcement Learning (ORL) is a promising approach to reduce the high sample complexity of traditional Reinforcement Learning (RL) by eliminating the need for continuous environmental interactions. ORL exploits a dataset of pre-collected transitions and thus expands the range of application of RL to tasks in which the excessive environment queries increase training time and decrease efficiency, such as in modern AAA games. This paper introduces OfflineMania a novel environment for ORL research. It is inspired by the iconic TrackMania series and developed using the Unity 3D game engine. The environment simulates a single-agent racing game in which the objective is to complete the track through optimal navigation.
We provide a variety of datasets to assess ORL performance. These datasets, created from policies of varying ability and in different sizes, aim to offer a challenging testbed for algorithm development and evaluation. We further establish a set of baselines for a range of Online RL, ORL, and hybrid Offline to Online RL approaches using our environment.

\end{abstract}

\section{Introduction}

Reinforcement Learning (RL) has become increasingly popular in the gaming industry as it offers a promising way of creating immersive gaming experiences. From training AI-controlled non-player characters~\citep{alpha,gt} to automated game testing~\citep{alessandrogt, gametesting}.

However the widespread use of RL is often limited by its sample complexity which makes training in complex environments, such as modern AAA games, slow and inefficient. Offline RL (ORL) has recently garnered interest as a framework aimed at improving the sample efficiency of RL agents~\citep{offline}. With ORL, one can completely eliminate the need for interaction with the environment and instead rely on a previously collected dataset of experiences. Such datasets can be made readily accessible to game developers; for instance, they could use samples obtained from playtesting sessions or data extracted from previously released games.

In this paper we introduce OfflineMania a novel game environment for Online RL and ORL, centered around a single-agent racing game inspired by the iconic \textit{TrackMania}~\citep{track} game series. Our environment, built using the Unity 3D game engine~\citep{unity}, provides a track in which the agent must complete the race through optimal navigation. Moreover, we provide datasets of agent experiences tailored specifically for benchmarking ORL techniques. These datasets are of varying quality, ranging from those generated using random policies to those crafted by expert agents. We additionally offer smaller and mixed versions of these datasets. These variants are designed to test algorithmic performance under complex scenarios that challenge the robustness of learning methods.
Our work focuses on providing a gaming testbed environment \emph{and} multiple datasets tailored for game AI research in ORL. To the best of our knowledge such a combination is not currently available in the existing literature.
We further provide a study assessing the performance of different Online RL and ORL algorithms using our new datasets. We also investigate the performance of fine-tuning policies trained offline using Online RL. This last framework is a more natural approach in game development, as it allows developers to take advantage of datasets to create a policy that then can be effectively improved with fewer game interactions.

\begin{figure}
    \centering
    \includegraphics[width=.4\textwidth]{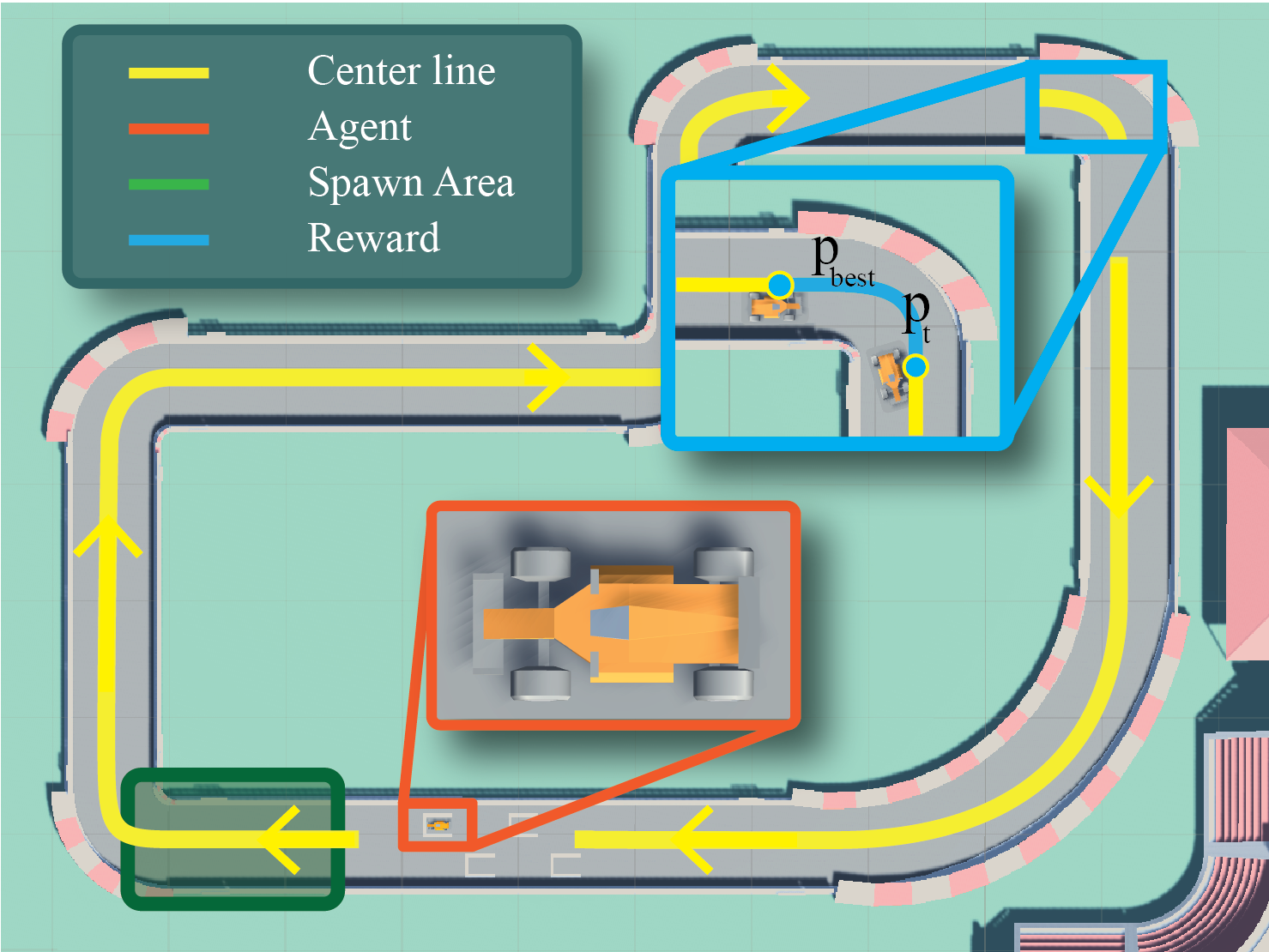}
    \caption{Visualization of OfflineMania, highlighting the track centerline used in the reward function, the episode starting area, the agent, and the positive part of the reward during a transition. \vspace{-0.4cm}
    }
    \label{fig:track}
\end{figure}

The key contributions of this work are:
\begin{itemize}
    \item we introduce OfflineMania, a new environment inspired by TrackMania developed using the Unity 3D engine; 

    \item we provide diverse datasets of varying sizes, collected using policies of different expertise levels; and
    
    \item we present results for a variety of baseline algorithms including both Online and ORL approaches, as well as hybrid methods that combine Offline training with Online fine-tuning. 
\end{itemize}

\section{Related work}

\minisection{Offline Reinforcement Learning (ORL).} Offline Reinforcement Learning is a promising way to address the sample complexity challenges faced by Online RL. The core objective of ORL is to create a robust policy from a fixed dataset of pre-collected environment transitions, without requiring further online interactions with the environment~\citep{IQL, TD3BC, cql}. ORL has emerged as a promising approach to training game agents~\citep{ubisoftoffline}. Modern AAA games are often computationally demanding, slow to simulate, and inherently unstable, all of which lead to increased need for extensive interactions with the environment.

ORL has significantly benefited from the development of benchmarks like D4RL~\citep{d4rl}, which provides a wide range of datasets across various domains such as locomotion, robotic manipulation, and vision-based autonomous driving. However, to the best of our knowledge, there is no ORL dataset available in the literature specifically tailored for studying these techniques in gaming environments, particularly in the context of racing games. This paper aims to bridge this gap.

\minisection{Offline to Online RL.} The Offline to Online approach focuses on how to effectively \emph{improve} policies trained with ORL in a online setting which allows further environment interactions. It is a promising approach for training game agents~\cite{kiwi} since a game in development can change on a daily basis, potentially rendering the datasets collected in an environment iteration insufficient for subsequent iterations. Offline to Online RL can allow game developers to seamlessly transition to new environment with minimal online interactions. However, this transition poses significant challenges~\citep{awac,lee}, including dealing with distribution shifts between the offline data and the new online interactions. For these reasons, with this paper we aim to provide a benchmark suite for investigating the impact of Offline to Online training in gaming environments.

\section{Environment and Datasets}

\subsection{Environment}
We developed OfflineMania, shown in Figure~\ref{fig:track}, using the Unity 3D game engine leveraging the ML-Agents package~\citep{unity}. It features a Gymnasium-compatible interface~\cite{towers_gymnasium_2023}, ensuring straightforward integration into existing experimental setups. The game is computationally efficient, with the game speed easily adjustable to speed up online training process. Additionally, the environment supports rendering capabilities, offering a birds-eye view of the car. This visualization facilitates qualitative assessment of agent behavior and simplifies the evaluation process. 

We now describe the main elements of the environment: the state and action-space, the reward function, and episode loop.

\minisection{State Space.} The state space is a vector in $\mathcal{R}^{33}$. It is composed by 15 raycasts covering a 180-degree field of view in front of the car, with each ray are associated two values: one indicating the presence of an object within its path, and the other specifying the distance to the detected object. Additionally, the components of the velocity of the car are included as part of the state representation. 
	
\minisection{Action Space.} The agent action space consists of two continuous values. The first value controls the steering angle of the car which ranges from -1 (indicating a left turn) to 1 (indicating a right turn). The second value controls the acceleration or braking of the car, with a value of 1 corresponding to full acceleration and a value of -1 representing braking or reversing when the car is stationary.


\minisection{Reward Signal.} Our reward function draws inspiration from prior work~\citep{gt}. We denote with $p_t$ the position of the car projected onto the track centerline at timestep $t$,
and by $p_{\text{best}}$ the most advanced position achieved thus far in the episode. Our reward is then:
\begin{equation}
    r_t = r_t^{\text{prog}} - 
    \begin{cases}
        \lambda \parallel v_{\text{car}}\parallel  & \text{if in contact with wall} \\
        0 & \text{otherwise}
    \end{cases},
\end{equation}
where $r_t^{\text{prog}}$ quantifies the progress of position $p_t$ along the centerline relative to $p_{\text{best}}$, as shown in Figure~\ref{fig:track}. We set $r_t^{\text{prog}}$ to 0 if $p_t$ does not advance beyond $p_{\text{best}}$.
$v_{\text{car}}$ is the magnitude of the velocity at the moment of impact, and $\lambda$ is a fixed coefficient penalizing collisions. In our environment, $\lambda = 50$.


\minisection{Episode.}
During each episode the vehicle starts with its position dynamically chosen within a designated square area before the fist turn in the track. We also randomize the orientation of the car, between -30 and 30 degrees from the centerline, which ensures that it is always facing the correct direction.
Every episode has a fixed length of 2,000 steps. In this many steps an expert agent can complete at most 5 laps.

\subsection{Datasets}
\label{sec:datasets}
In order to support comprehensive research in ORL, we generated a diverse series datasets. First, we train three distinct policies using Proximal Policy Optimization (PPO)~\citep{PPO}, stopping the training after 1,000, 5,000, and 12,500 network updates, respectively. Each of the trained policy represents varying degrees of ability in navigating the race track. The policies obtain mean cumulative reward of -360, 327, and 1183 respectively, over five episodes. The first policy struggles with the initial corner. The second policy, while capable of occasionally completing the track, exhibits inconsistent performance. In contrast, the third policy consistently achieved high performance by efficiently navigating the track, including corner-cutting strategies, successfully completing 5 laps in each episode.

Using these three policies, we collected three distinct datasets: \textit{basic}, \textit{medium}, and \textit{expert}, each consisting of 100,000 transitions. Additionally, we created mixed datasets, one consisting of 200,000 transitions in total and another of only 5,000. We refer to these datasets with \textit{mix large} and \textit{mix small}, respectively. The mixed datasets consist of 90\% transitions sampled from the basic policy, 7\% from the medium policy, and only 3\% from the expert policy. The distribution of transitions in the mixed datasets was chosen to simulate a complex scenario in which ORL algorithms must stitch together different behaviors in order to correctly learn an optimal policy. The smaller version of the mixed dataset is useful to understand the behavior of ORL approaches when dealing with small datasets. Following this idea, we built another datasets from only 5,000 transitions collected exclusively from the basic policy, called \textit{basic small}. Those two variants -- \textit{mix small} and \textit{basic small} -- offer a more demanding testbed for evaluating the robustness and adaptability of ORL algorithms under complex training conditions.

\section{Benchmark Study}
\label{sec:experiments}
OfflineMania aims to be a testbed for developing new training techniques. We provide results of a set of baselines for widely recognized methods for Online RL, ORL, and Offline to Online RL. For all approaches we present the mean results over five different seeds. All experiments were conducted using a system equipped with a Nvidia RTX 2070 and an AMD Ryzen 3600X processor.

For Online RL baselines, we opted for two state-of-the-art methods: Proximal Policy Optimization (PPO)~\citep{PPO} and Soft Actor Critic (SAC)~\citep{SAC}. PPO, known for its efficacy and robustness to hyperparameter selection, represents a widely used policy-based approach. Similarly, we selected SAC, an actor-critic algorithm, for its efficiency and for its importance in the RL landscape. We trained our PPO agent over 12,500 network updates, for a total of 15 million environment interactions and approximately 10 hours of training time. In contrast, training with SAC spanned 3 million network updates, with an equivalent number of environment interactions, totaling approximately 20 hours of training time.

For ORL approaches, we chose Conservative Q-Learning (CQL)~\citep{cql}, Twin Delayed Deep Deterministic policy gradient with Behavioral cloning (TD3BC)~\citep{TD3BC}, and Implicit Q-Learning (IQL)~\citep{IQL}. For all algorithms we present results after 300,000 network updates, corresponding to about one hour of training time for all algorithms.

For Offline to Online approaches, we compare various methods. These include: an approach combining TD3BC~\citep{TD3BC} for offline training and TD3~\citep{TD3} for online fine tuning; IQL~\citep{IQL}, following the fine-tuning process outlined in the original paper; Jump Start Reinforcement Learning (JSRL)~\citep{jsrl}, which utilizes offline policies as guides for online training; Policy EXpainsion (PEX)~\citep{pex}, a method combining offline policies with online training to enhance exploration; and the work of \citet{SDBG} that we will refer as SDBG, designed specifically for small offline datasets, utilizing a world model based augmentation to improve offline training. For each approach we present results after 300,000 offline network updates and 1 million online fine-tuning updates, for a total of about 4 hours of training across all algorithms. 
Since PEX, SDGB, and JSRL are agnostic to the algorithm used, we decided to show results using IQL.


\begin{table}
\centering
\caption{Average Rewards of Offline Reinforcement Learning Training  after 300,000 Network Updates.} 
\label{offline_res}
\begin{tabular}{lrrr} \toprule
Methods & \thead{TD3BC}       & \thead{CQL}   & \thead{IQL}           \\ \midrule
Expert      & -3981$\pm$57\phantom{0}    & -3325$\pm$1353 & \textbf{1192$\pm$1\phantom{00}} \\
Medium      & 335$\pm$87\phantom{0}        & -4227$\pm$571\phantom{0} & \textbf{789$\pm$58\phantom{0}}  \\
Basic       & 12$\pm$9\phantom{00}     & 39$\pm$81\phantom{00} & \textbf{98$\pm$38\phantom{0}}   \\
Mix Large   & 219$\pm$96\phantom{0}         & -4080$\pm$873\phantom{0} & \textbf{828$\pm$38\phantom{0}}  \\
Mix Small   & -1125$\pm$154       & -3972$\pm$858\phantom{0} & \textbf{10$\pm$32\phantom{0}}   \\
Basic Small & \textbf{64$\pm$31\phantom{0}} & -3488$\pm$626\phantom{0} & 20$\pm$102    \\   \bottomrule

\end{tabular}
\end{table}
\subsection{Online RL Results}
We use online RL for training a policy from scratch with environment interactions. The resulting mean reward achieved by PPO at the end of the training was 1183, indicative of consistent high-quality behaviors on the track. The policy is proficient at navigating the track, and is also able to cut corners effectively. It can complete a lap in 385 steps, for a total of 5 laps in a single episode. PPO succeeds at the cost of a large number of environment interactions (about 15 million).

Conversely, despite training for 20 hours and 3 million environment interactions, the SAC outcomes are considerably less promising. The SAC policy achieves a total mean reward of only 215 and demonstrates suboptimal performance characterized by slower and less stable navigation of the track compared to the PPO-trained policy. This highlights the limitations of SAC, despite its better sample efficiency, in producing effective policies even after extensive training.

These results quantify the efficacy of Online RL approaches. However, they also highlight the challenges posed by the slow training speed and significant sample inefficiency.

\subsection{Offline RL Results}
In Table~\ref{offline_res} we present the results for ORL baselines on the datasets described in Section~\ref{sec:datasets}. For all algorithms the training time for 300,000 network updates was under one hour. Remarkably, IQL consistently outperforms TD3BC and CQL across nearly all datasets. Of particular note is the ability of IQL to learn policies from the expert dataset that even surpass the performance of the policy used to generate it. 
On all other datasets, IQL produces policies that are capable of navigating the track without major collisions.

Conversely, TD3BC and CQL unexpectedly fall short on all datasets, most notably on the expert dataset. This may be attributed to sensitivity to the hyperparameters that is a well known problem in ORL~\citep{offline}.





\begin{table*}
\centering
\caption{Average Rewards for Offline-to-Online Approaches after 300,000 Offline Network Updates and 1 Million Online Updates.}
\label{tab:off_online_results}
\begin{tabular}{lrrrrr} \toprule
Methods & \thead{TD3BC+TD3}      & \thead{IQL}           & \thead{SDBG-IQL}                                 & \thead{PEX-IQL} & \thead{JSRL-IQL} \\ \midrule
Expert      & 15$\pm$11\phantom{00}    & 1192$\pm$4\phantom{00}          &\textbf{1199$\pm$3\phantom{00}}   & -2425$\pm$2132   & 1193$\pm$1\phantom{0}     \\
Medium      & -497$\pm$773\phantom{0}   & 856$\pm$58\phantom{0}  &  852$\pm$12\phantom{0}                   & 776$\pm$39\phantom{00}     & \textbf{856$\pm$1\phantom{0}}\\
Basic       & 11$\pm$7\phantom{000}       & 71$\pm$13\phantom{0}           &\textbf{130$\pm$35\phantom{0}} & -173$\pm$504\phantom{0}    & 99$\pm$33       \\
Mix Large   & -2538$\pm$1170  & 932$\pm$50\phantom{0}          & \textbf{1152$\pm$12\phantom{0}}            & 671$\pm$175\phantom{0}     & 1142$\pm$2\phantom{0}     \\
Mix Small   & -3886$\pm$923\phantom{0}   & -614$\pm$882          & \textbf{340$\pm$187}                     & -2147$\pm$1561   & -826$\pm$86     \\
Basic Small & -4352$\pm$1425   & 3$\pm$33\phantom{0}             & 105$\pm$3\phantom{00}                    & -2200$\pm$2042   & \textbf{152$\pm$1\phantom{0}}\\ \bottomrule
\end{tabular}
\end{table*}

        
            
                 
            
            


\subsection{Offline to Online RL Results}
In this section we report baseline performance for the combination of ORL pre-training and Online RL fine-tuning. Such approaches are useful when an offline policy fails to meet deployment standards. In these cases, we would like to improve the offline-trained policy by allowing some interactions with the environment, while still minimizing such interactions.

While this process might appear straightforward, it is complex primarily due to the distributional shift problem. Distributional shift occurs during the first training iterations when moving to online training, as the agent navigates into unexplored state-action spaces. In this setting the values of the Q-function trained during the offline phase may become highly inaccurate. This inaccuracy can lead to incorrect policy evaluations and arbitrary policy updates in these unseen states, which undermines the policy learned through ORL~\citep{lee}. 

In Table~\ref{tab:off_online_results} we present the results of the experiments in this setup. The combination of TD3BC and TD3 fails to improve the offline policy and achieves worse performances on all datasets. This might be due to the distributional shift problem which is not explicitly addressed by this approach. 
On the other hand, IQL demonstrates good fine-tuning performance and improves its offline results in almost all datasets. Only for the \textit{mix small} dataset the score is substantially reduced. This is probably due to the difficulty in learning with such a small dataset derived from different policies.


Although PEX attempts to directly address the challenge of training an online policy after offline pre-training, it achieves the worst performance -- lower even than the standard fine-tuning process of IQL. On the other hand, SDBG and JSRL show remarkable performance. Even though SDBG specifically addresses training setups with small datasets -- as seen in the \textit{mix small} and \textit{basic small} results from Table~\ref{tab:off_online_results} -- it is the only algorithm that improves the offline performance in \emph{all} tasks. JSRL achieves good performance across all tasks as well, except again for the \textit{mix small} dataset.

\section{Discussion}
ORL is a promising approach to fostering more widespread use of RL in modern video games, however there is a lack of environments with datasets that can be used to test the capabilities of ORL methods in challenging conditions. This paper introduces a novel game environment which serves as a benchmark for ORL research. We provide datasets of varying data quality and quantity which facilitates the simulation of complex training scenarios\footnote{Env. and datasets are available at: \url{https://github.com/ganjiro/OfflineMania}}. Additionally, we present results from state-of-the-art methods for Online RL, ORL, and hybrid Offline to Online RL approaches. This benchmark aims to facilitate future investigation into the use of offline data in gaming environments. 
By leveraging offline datasets, we can effectively mitigate the challenges in applying RL techniques to modern games, thereby contributing the integration of RL into modern game development workflows.

\bibliographystyle{IEEEtranN}
{\footnotesize \bibliography{biblio}}

\begin{thebibliography}{22}
\providecommand{\natexlab}[1]{#1}
\providecommand{\url}[1]{#1}
\csname url@samestyle\endcsname
\providecommand{\newblock}{\relax}
\providecommand{\bibinfo}[2]{#2}
\providecommand{\BIBentrySTDinterwordspacing}{\spaceskip=0pt\relax}
\providecommand{\BIBentryALTinterwordstretchfactor}{4}
\providecommand{\BIBentryALTinterwordspacing}{\spaceskip=\fontdimen2\font plus
\BIBentryALTinterwordstretchfactor\fontdimen3\font minus \fontdimen4\font\relax}
\providecommand{\BIBforeignlanguage}[2]{{%
\expandafter\ifx\csname l@#1\endcsname\relax
\typeout{** WARNING: IEEEtranN.bst: No hyphenation pattern has been}%
\typeout{** loaded for the language `#1'. Using the pattern for}%
\typeout{** the default language instead.}%
\else
\language=\csname l@#1\endcsname
\fi
#2}}
\providecommand{\BIBdecl}{\relax}
\BIBdecl

\bibitem[Silver et~al.(2016)Silver, Huang, Maddison, Guez, Sifre, Van Den~Driessche, Schrittwieser, Antonoglou, Panneershelvam, Lanctot, et~al.]{alpha}
D.~Silver, A.~Huang, C.~J. Maddison, A.~Guez, L.~Sifre, G.~Van Den~Driessche, J.~Schrittwieser, I.~Antonoglou, V.~Panneershelvam, M.~Lanctot \emph{et~al.}, ``Mastering the game of go with deep neural networks and tree search,'' \emph{nature}, 2016.

\bibitem[Fuchs et~al.(2021)Fuchs, Song, Kaufmann, Scaramuzza, and D{\"u}rr]{gt}
F.~Fuchs, Y.~Song, E.~Kaufmann, D.~Scaramuzza, and P.~D{\"u}rr, ``Super-human performance in gran turismo sport using deep reinforcement learning,'' \emph{IEEE Robotics and Automation Letters}, 2021.

\bibitem[Sestini et~al.(2022)Sestini, Gisslén, Bergdahl, Tollmar, and Bagdanov]{alessandrogt}
A.~Sestini, L.~Gisslén, J.~Bergdahl, K.~Tollmar, and A.~D. Bagdanov, ``Automated gameplay testing and validation with curiosity-conditioned proximal trajectories,'' \emph{IEEE Transactions on Games}, 2022.

\bibitem[Bergdahl et~al.(2020)Bergdahl, Gordillo, Tollmar, and Gissl{\'e}n]{gametesting}
J.~Bergdahl, C.~Gordillo, K.~Tollmar, and L.~Gissl{\'e}n, ``Augmenting automated game testing with deep reinforcement learning,'' in \emph{IEEE Conference on Games (CoG)}, 2020.

\bibitem[Levine et~al.(2020)Levine, Kumar, Tucker, and Fu]{offline}
S.~Levine, A.~Kumar, G.~Tucker, and J.~Fu, ``Offline reinforcement learning: Tutorial, review, and perspectives on open problems,'' \emph{arXiv preprint arXiv:2005.01643}, 2020.

\bibitem[Ubisoft(2020)]{track}
\BIBentryALTinterwordspacing
Ubisoft, ``Trackmania,'' 2020. [Online]. Available: \url{https://www.ubisoft.com/en-us/game/trackmania/trackmania}
\BIBentrySTDinterwordspacing

\bibitem[Juliani et~al.(2018)Juliani, Berges, Teng, Cohen, Harper, Elion, Goy, Gao, Henry, Mattar, et~al.]{unity}
A.~Juliani, V.-P. Berges, E.~Teng, A.~Cohen, J.~Harper, C.~Elion, C.~Goy, Y.~Gao, H.~Henry, M.~Mattar \emph{et~al.}, ``Unity: A general platform for intelligent agents,'' \emph{arXiv preprint arXiv:1809.02627}, 2018.

\bibitem[Kostrikov et~al.(2021)Kostrikov, Nair, and Levine]{IQL}
I.~Kostrikov, A.~Nair, and S.~Levine, ``Offline reinforcement learning with implicit q-learning,'' \emph{arXiv preprint arXiv:2110.06169}, 2021.

\bibitem[Fujimoto and Gu(2021)]{TD3BC}
S.~Fujimoto and S.~S. Gu, ``A minimalist approach to offline reinforcement learning,'' \emph{Advances in neural information processing systems}, 2021.

\bibitem[Kumar et~al.(2020)Kumar, Zhou, Tucker, and Levine]{cql}
A.~Kumar, A.~Zhou, G.~Tucker, and S.~Levine, ``Conservative q-learning for offline reinforcement learning,'' \emph{Advances in Neural Information Processing Systems}, 2020.

\bibitem[Kobanda et~al.(2023)Kobanda, Valliappan, Romoff, and Denoyer]{ubisoftoffline}
A.~Kobanda, C.~Valliappan, J.~Romoff, and L.~Denoyer, ``Learning computational efficient bots with costly features,'' in \emph{IEEE Conference on Games (CoG)}, 2023.

\bibitem[Fu et~al.(2020)Fu, Kumar, Nachum, Tucker, and Levine]{d4rl}
J.~Fu, A.~Kumar, O.~Nachum, G.~Tucker, and S.~Levine, ``D4rl: Datasets for deep data-driven reinforcement learning,'' \emph{arXiv preprint arXiv:2004.07219}, 2020.

\bibitem[Sestini et~al.(2023)Sestini, Bergdahl, Tollmar, Bagdanov, and Gissl{\'e}n]{kiwi}
A.~Sestini, J.~Bergdahl, K.~Tollmar, A.~D. Bagdanov, and L.~Gissl{\'e}n, ``Towards informed design and validation assistance in computer games using imitation learning,'' in \emph{IEEE Conference on Games (CoG)}, 2023.

\bibitem[Nair et~al.(2020)Nair, Gupta, Dalal, and Levine]{awac}
A.~Nair, A.~Gupta, M.~Dalal, and S.~Levine, ``Awac: Accelerating online reinforcement learning with offline datasets,'' \emph{arXiv preprint arXiv:2006.09359}, 2020.

\bibitem[Lee et~al.(2022)Lee, Seo, Lee, Abbeel, and Shin]{lee}
S.~Lee, Y.~Seo, K.~Lee, P.~Abbeel, and J.~Shin, ``Offline-to-online reinforcement learning via balanced replay and pessimistic q-ensemble,'' in \emph{Conference on Robot Learning}.\hskip 1em plus 0.5em minus 0.4em\relax PMLR, 2022, pp. 1702--1712.

\bibitem[Towers et~al.(2023)Towers, Terry, Kwiatkowski, Balis, Cola, Deleu, Goulão, Kallinteris, KG, et~al.]{towers_gymnasium_2023}
\BIBentryALTinterwordspacing
M.~Towers, J.~K. Terry, A.~Kwiatkowski, J.~U. Balis, G.~d. Cola, T.~Deleu, M.~Goulão, A.~Kallinteris, A.~KG \emph{et~al.}, ``Gymnasium,'' Mar. 2023. [Online]. Available: \url{https://zenodo.org/record/8127025}
\BIBentrySTDinterwordspacing

\bibitem[Schulman et~al.(2017)Schulman, Wolski, Dhariwal, Radford, and Klimov]{PPO}
J.~Schulman, F.~Wolski, P.~Dhariwal, A.~Radford, and O.~Klimov, ``Proximal policy optimization algorithms,'' \emph{arXiv preprint arXiv:1707.06347}, 2017.

\bibitem[Haarnoja et~al.(2018)Haarnoja, Zhou, Abbeel, and Levine]{SAC}
T.~Haarnoja, A.~Zhou, P.~Abbeel, and S.~Levine, ``Soft actor-critic: Off-policy maximum entropy deep reinforcement learning with a stochastic actor,'' in \emph{International conference on machine learning}.\hskip 1em plus 0.5em minus 0.4em\relax PMLR, 2018.

\bibitem[Fujimoto et~al.(2018)Fujimoto, Hoof, and Meger]{TD3}
S.~Fujimoto, H.~Hoof, and D.~Meger, ``Addressing function approximation error in actor-critic methods,'' in \emph{International conference on machine learning}.\hskip 1em plus 0.5em minus 0.4em\relax PMLR, 2018.

\bibitem[Uchendu et~al.(2023)Uchendu, Xiao, Lu, Zhu, Yan, Simon, Bennice, Fu, Ma, Jiao, et~al.]{jsrl}
I.~Uchendu, T.~Xiao, Y.~Lu, B.~Zhu, M.~Yan, J.~Simon, M.~Bennice, C.~Fu, C.~Ma, J.~Jiao \emph{et~al.}, ``Jump-start reinforcement learning,'' in \emph{International Conference on Machine Learning}.\hskip 1em plus 0.5em minus 0.4em\relax PMLR, 2023.

\bibitem[Zhang et~al.(2023)Zhang, Xu, and Yu]{pex}
H.~Zhang, W.~Xu, and H.~Yu, ``Policy expansion for bridging offline-to-online reinforcement learning,'' \emph{arXiv preprint arXiv:2302.00935}, 2023.

\bibitem[Macaluso et~al.(2024)Macaluso, Sestini, and Bagdanov]{SDBG}
G.~Macaluso, A.~Sestini, and A.~D. Bagdanov, ``Small dataset, big gains: Enhancing reinforcement learning by offline pre-training with model-based augmentation,'' in \emph{Computer Sciences \& Mathematics Forum}.\hskip 1em plus 0.5em minus 0.4em\relax MDPI, 2024.

\end{thebibliography}

\end{document}